\documentclass[10pt,twocolumn,letterpaper]{article}

\usepackage{cvpr}
\usepackage{times}
\usepackage{epsfig}
\usepackage{graphicx}
\usepackage{amsmath}
\usepackage{amssymb}
\usepackage{xcolor}
\usepackage{tabu,multirow}
\usepackage{url}
\usepackage[mathcal]{euscript}
\usepackage{flushend}

\usepackage[pagebackref=true,breaklinks=true,letterpaper=true,colorlinks,bookmarks=false]{hyperref}

\cvprfinalcopy 

\ifcvprfinal\fi
\begin{document}
\newcommand{\mbf}[1]{\ensuremath{\mathbf{#1}}}
\newcommand{\mbs}[1]{\ensuremath{\boldsymbol{#1}}}
\newcommand{\mcl}[1]{\ensuremath{\mathcal{#1}}}
\newcommand{\mrm}[1]{\ensuremath{\mathrm{#1}}}
\newcommand{\mbb}[1]{\ensuremath{\mathbb{#1}}}
\newcommand{\msf}[1]{\ensuremath{\mathsf{#1}}}

\newcommand{\ve}[1]{\ensuremath{\mathbf{#1}}} 
\newcommand{\m}[1]{\ensuremath{\mathsf{#1}}} 
\newcommand{\s}[1]{\ensuremath{\mathcal{#1}}} 

\newcommand{\trsp}[1]{\ensuremath{#1^{\top}}}
\newcommand{\pinv}[1]{\ensuremath{#1^{\dagger}}}
\newcommand{\bmat}[4]{\ensuremath{\begin{bmatrix}#1&#2\\#3&#4\end{bmatrix}}}
\def\trace{\ensuremath{\mathrm{trace}}}
\def\deter{\ensuremath{\mathrm{det}}}
\def\diag{\ensuremath{\mathrm{diag}}}
\def\rank{\ensuremath{\mathrm{rank}}}
\def\Id{\m{Id}} 
\def\mA{\m{A}}
\def\mB{\m{B}}
\def\mC{\m{C}}
\def\mD{\m{D}}
\def\mE{\m{E}}
\def\mF{\m{F}}
\def\mG{\m{G}}
\def\mH{\m{H}}
\def\mK{\m{K}}
\def\mL{\m{L}}
\def\mN{\m{M}}
\def\mP{\m{P}}
\def\mW{\m{W}}
\def\mX{\m{X}}
\def\mY{\m{Y}}
\def\mZ{\m{Z}}

\newcommand{\bvec}[2]{\ensuremath{\begin{bmatrix}#1\\#2\end{bmatrix}}}
\def\One{\mbs{1}} 
\def\va{\ve{a}}
\def\vb{\ve{b}}
\def\vc{\ve{c}}
\def\vd{\ve{d}}
\def\vf{\ve{f}}
\def\vg{\ve{g}}
\def\vh{\ve{h}}
\def\vi{\ve{i}}
\def\vt{\ve{t}}
\def\bx{\ve{x}}
\def\by{\ve{y}}
\def\bz{\ve{z}}

\def\cL{\mcl{L}}

\def\ie{\emph{i.e.}}
\def\eg{\emph{e.g.}}
\def\iid{\emph{i.i.d.}}
\def\wrt{w.r.t.}
\def\mwrt{\mrm{w.r.t.}}
\def\msbt{\mrm{sb.t.}}

\def\sqt{^{\frac{1}{2}}} 
\def\msqt{^{-\frac{1}{2}}} 
\def\R{\mbb R}
\def\vtheta{\mbs{\theta}}

\title{Structural inpainting}

\author{Huy V. Vo, Ngoc Q. K. Duong, Patrick P\'erez\\
Technicolor\\
{\tt\small van-huy.vo@polytechnique.edu,Quang-Khanh-Ngoc.Duong@technicolor.com,patrick.perez@valeo.com}\\~\\
}

\twocolumn[{%
	\renewcommand\twocolumn[1][]{#1}%
	\maketitle
	{
    	\centering
		\vspace{-0.8cm}
		\includegraphics[width=\linewidth]{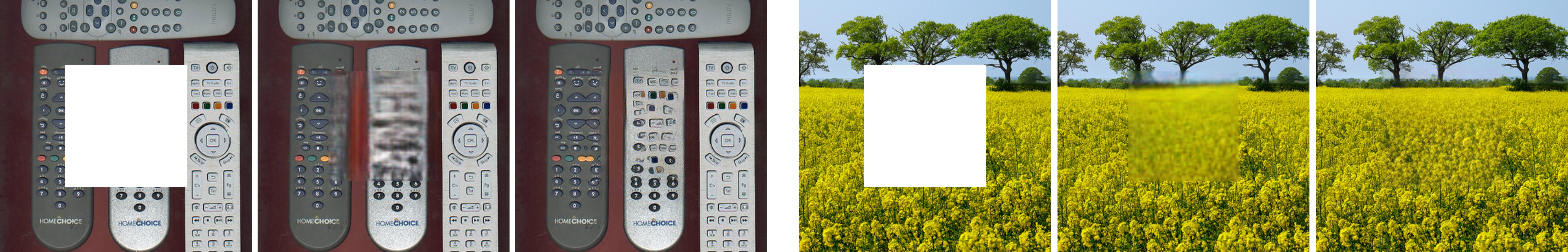}
	}\\
	Given an arbitrary incomplete scene (left image), the proposed context encoder produces a plausible structural completion (middle image), which can be subsequently refined for texture and details with a patch-based inpainting method (right image).
	\label{fig:teaser}
	\vspace{0.5cm}
}]

\begin{abstract}
Scene-agnostic visual inpainting remains very challenging despite progress in patch-based methods. Recently, Pathak \etal \cite{pathak2016context} have introduced convolutional ``context encoders'' (CEs) for unsupervised feature learning through image completion tasks. With the additional help of adversarial training, CEs turned out to be a promising tool to complete complex structures in real inpainting problems. In the present paper we propose to push further this key ability by relying on perceptual reconstruction losses at training time. We show on a wide variety of visual scenes the merit of the approach for \textit{structural inpainting}, and confirm it through a user study. Combined with the optimization-based refinement of \cite{yang2016high} with neural patches, our context encoder opens up new opportunities for prior-free visual inpainting.     
\end{abstract}

\section{Introduction}

Visual inpainting \cite{bertalmio2000image}, a.k.a. scene completion, is the task of filling in a plausible way a region in an image. Automatic and semi-automatic inpainting tools are important for both restoration and editing of visual content, whether photographs or video sequences. Professionals and amateurs use them to replace missing or damaged fragments of scanned content, and to remove undesirable scene elements from small imperfections to complete occluding objects. 

In the general case where the target region can be of any relative size and the content of the scene can be of any type, most successful approaches are based on ``patches'': The target region is filled in, either in a greedy fashion \cite{criminisi2003object,drori2003fragment} or by iterative optimization \cite{wexler2004space,barnes2009patchmatch}, using image blocks selected in the surrounding. In the absence of strong prior on the scene or restrictions on the hole's dimensions, such methods are indeed the only up to now that can generate plausible structures and textures. 

However, with the rapid progress of deep convolutional neural nets, inpainting has been revisited recently as yet another visual prediction problem for which a deep regressor can be learned \cite{pathak2016context}. Pathak \etal's \textit{context encoder} (CE) is trained on diverse images to infer the (known) central part of the scene given its periphery, using a classic $\ell_2$ reconstruction loss along with an adversarial one. On new scenes, it produces very interesting completions with structures, if not textures, that are challenging for patch-based methods. While the final results still lack details, their quality can be improved by a patch-based refinement: Yang \etal \cite{yang2016high} use the output of Pathak's context encoder to guide an optimization-based inpainting where patches are described and compared using deep features. Impressive results are reported with this two-stage approach. 

Inspired by these recent developments, we analyse here some of the limits of context encoders and suggest simple, yet fruitful, modifications to the original design. We argue in particular that the ability of context encoders to complete scene structures is not as semantic as originally suggested and, more importantly, that CEs might under-perform even in not so complex situations. In order to strengthen this key ability to handle interrupted structures, we propose to rely on the perceptual losses that have recently proved useful for several other image modification tasks \cite{johnson2016perceptual}. By comparing ground-truth and predicted visual content in a deep feature domain rather than in the pixel domain when training, a perceptual loss provides more freedom to the regressor while focussing on meaningful image properties. We show that context encoders can also benefit from this powerful idea, which leads to improved \textit{structural inpainting}. With this modification of the original context encoder, adversarial training can be mobilized only in a second curriculum phase, to produce more realistic textures and details. As its predecessor, the new context encoder can finally be refined through a neural patch-based optimization \cite{yang2016high}. Results on a wide variety of scenes demonstrate the merit of the proposed approach, which is confirmed by a user study.     


\section{Problem and related work}

\begin{figure}
\includegraphics[width=\linewidth]{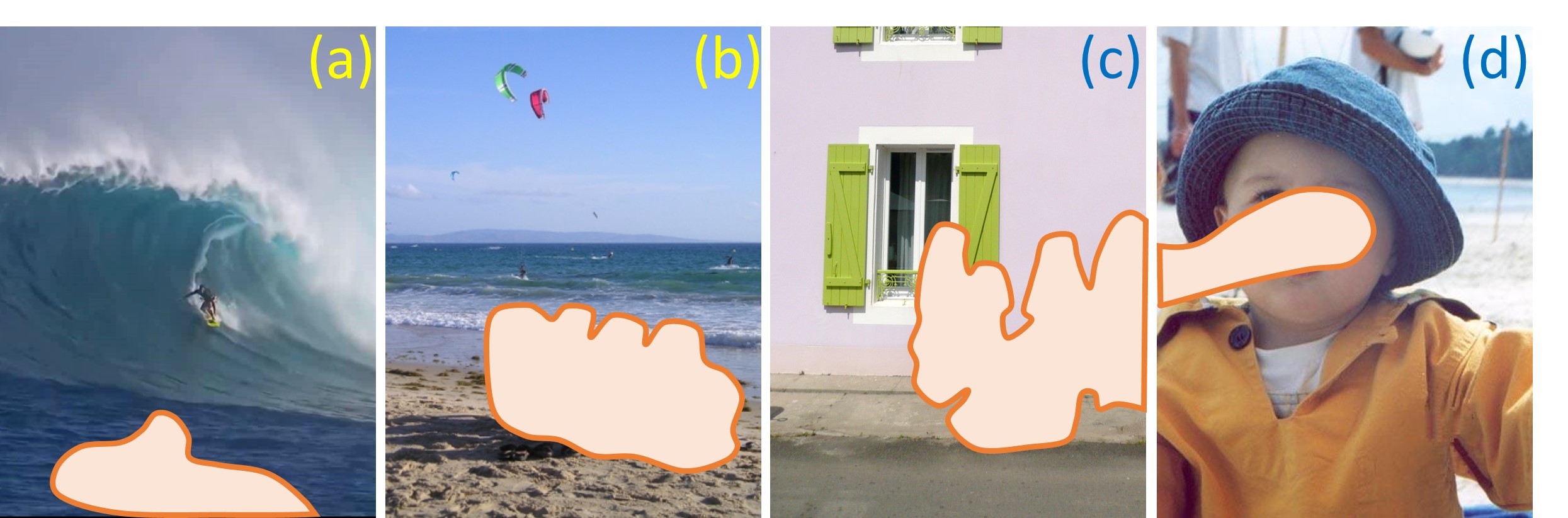}
\caption{\textbf{Generic inpainting problem}: How to complete plausibly the hole in the image, typically to remove an unwanted scene element? Several typical completion cases: (a) Single texture, many satisfactory fillings exist; (b) Multiple textures, the interface between the textured regions restricts reconstruction freedom; (c) Single or multiple structures, filling-in is very contrived; (d) Content with strong semantics, the most challenging case.}
\label{fig:inpainting}
\end{figure}

Visual inpainting comes in a variety of forms and names (completion, reconstruction, disocclusion, hallucination, recovery). Letting aside the specific problem of reconstructing an image from a fraction of its pixels -- which includes super-resolution and up-sampling -- a general formulation of the task is as follows: Given an image and a ``hole'' in it, that is a connected region of missing or unwanted content (Fig.\ \ref{fig:inpainting}), compute pixel values inside the hole such that the completed image looks natural or at least plausible. 

Generic inpainting tools, such as those present in image editing software packages, rely neither on prior knowledge about the scene nor on additional images (other views of the same scene or images of  similar scenes). The first successful approaches to such generic inpainting were geometric, relying on level line completion \cite{masnou1998level} and diffusion PDEs \cite{bertalmio2000image}. However, these techniques alone proved unable to handle large regions and to generate plausible texture. A major breakthrough came from leveraging non-parametric patch-based texture synthesis \cite{efros1999texture,bornard2002missing}.            

Initial patch-based inpainting lead to a large number of approaches, see \cite{buyssens2015exemplar} for a review. Two main families have emerged: Greedy approaches, \eg, \cite{criminisi2004region,drori2003fragment}, which fill the hole in a single concentric pass by copying patches from outside; Iterative optimization-based approaches, \eg, \cite{barnes2009patchmatch,wexler2004space,arias2011variational}, which alternate between computation of patch correspondences and image update through combination of matches. Both proved useful not only to handle texture well, but also to complete geometric structures to some extent. To further improve latter ability, specific mechanisms were proposed, \eg,  \cite{criminisi2003object,bugeau2010comprehensive,cao2011geometrically}. Another key aspect of patch-based methods is the way patches are matched: First using exhaustive or approximate pixel-wise color comparisons, later adding texture-aware features, \eg, \cite{liu2013exemplar}. More recently, ``neural patches'' have been used \cite{yang2016high}: Image patches are described using deep features provided by a convnet pre-trained for recognition.   

Yet, in case of too complex structures, like in indoor or urban scenes, and/or of too semantic content, like part of a face or a body, agnostic patch-based approaches fail catastrophically. For instance the example in Fig.\ \ref{fig:inpainting}(b) would be at best difficult for such approaches, and those in Figs.\ \ref{fig:inpainting}(c-d) completely infeasible. 

For such difficult but very common inpainting examples, several sources of help can come to the rescue: (1) User input, either through alternation of automatic processing and manual edits, or through scribbled indications \cite{sun2005image}; (2) Other relevant images, either other views of the exact same scene (in particular in the case of video inpainting \cite{wexler2004space,granados2012not,newson2014video}) or images of resembling scenes found in a large collection \cite{hays2007scene}; (3) Prior knowledge on the type of scene of interest, leading to class-specific inpainting.

Class-specific inpainting can be very powerful, but its scope is limited by essence. It requires the training of a class-specific appearance model that is then fitted to the visible part of the incoming image in order to infer the missing part. With a strong focus on faces, approaches of this type have been proposed using low dimensional models \cite{wang2007reconstruction}, sparse models \cite{burgos2015pose,sulam2016large} and, more recently, generative deep models \cite{yeh2017semantic,li2017generative}. 
 
In the present work, we are rather interested in scene-agnostic inpainting, for which major progress is still needed in a wide range of real use-cases, particularly regarding structure reconstruction. The recent deep learning approach of Pathak \etal \cite{pathak2016context} opens up new opportunities on that front. Our work builds on it. In the next section, we discuss this approach into more detail and investigate its current limitations.          


\section{Context encoders and their limits}

Pathak \etal \cite{pathak2016context} introduced the concept of context encoder, a deep encoder-decoder architecture trained to reconstruct images with missing parts. 
This self-supervised architecture is shown to produce appealing visual features for other tasks such as recognition, detection and semantic segmentation. In addition, it lends itself to actual scene completion. To this end, it is specifically trained to predict only the central part of an image given its periphery and the $\ell_2$ pixel-wise reconstruction loss is combined with an adversarial loss, which improves the visual quality of reconstructed regions. Trained on ImageNet ILSVRC'12 dataset, this context encoder yields impressive results on held-out images, while being scene-agnostic (or at least not specific to a single scene class). Its ability to recover complex, semantic structures is impressive in some cases where patch-based approaches are useless. When further specialized on urban scenes through training on Paris Street View dataset, this ability is increased accordingly on images of the same type. 

\begin{figure}[tb]
\includegraphics[width=1.0\linewidth]{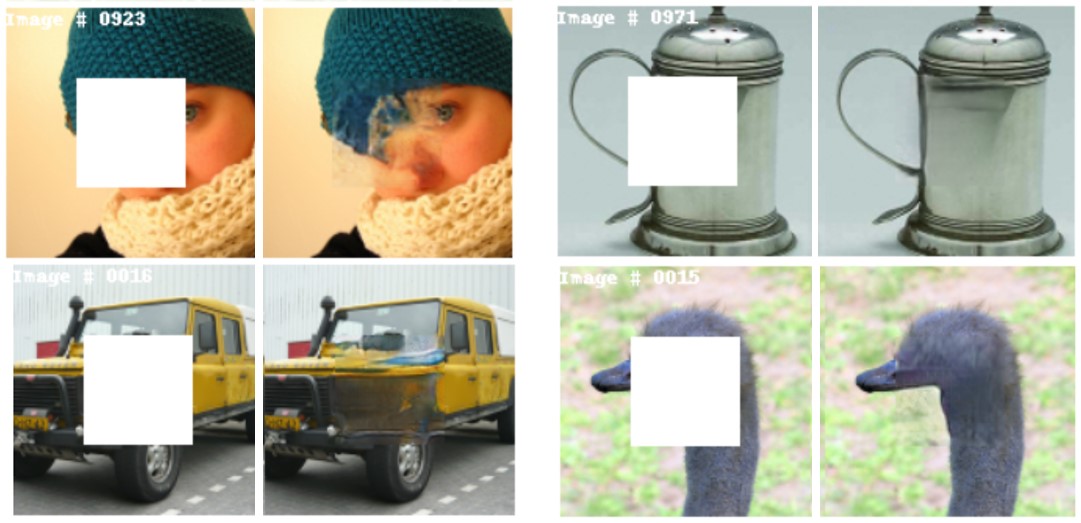}
\caption{\textbf{Scene completion by context encoding}: Failures (left) and encouraging results (right) when completing complex semantic structures with \cite{pathak2016context} (samples from \url{http://tinyurl.com/y8wv2zjy}).
}
\label{fig:bad_good}
\end{figure}

However, despite being said to allow ``semantic inpainting'', the CE trained on the wide range of object images in ImageNet often fails to produce plausible reconstructions of complex objects, such as humans, animals or cars (Fig.\ \ref{fig:bad_good}, left). This is not surprising since the CE is likely to capture only little class-specific structural knowledge, unless being specifically trained on a single object or scene class. In addition, as we shall see in our experiments, the surrounding context that CEs actually exploit is mostly local, sometimes only a few pixel wide with no access to visual semantics. 
Still, \cite{pathak2016context} reports several inpainting results where the proposed CE does much better than state-of-the-art patch-based methods on complex structures (Fig.\ \ref{fig:bad_good}, right). We believe this is where the main strength of context encoders lies, in addition to being optimization-free hence immune to initialization problems and potentially much faster. 

\begin{figure}[tb]
\centering\includegraphics[width=1.0\linewidth]{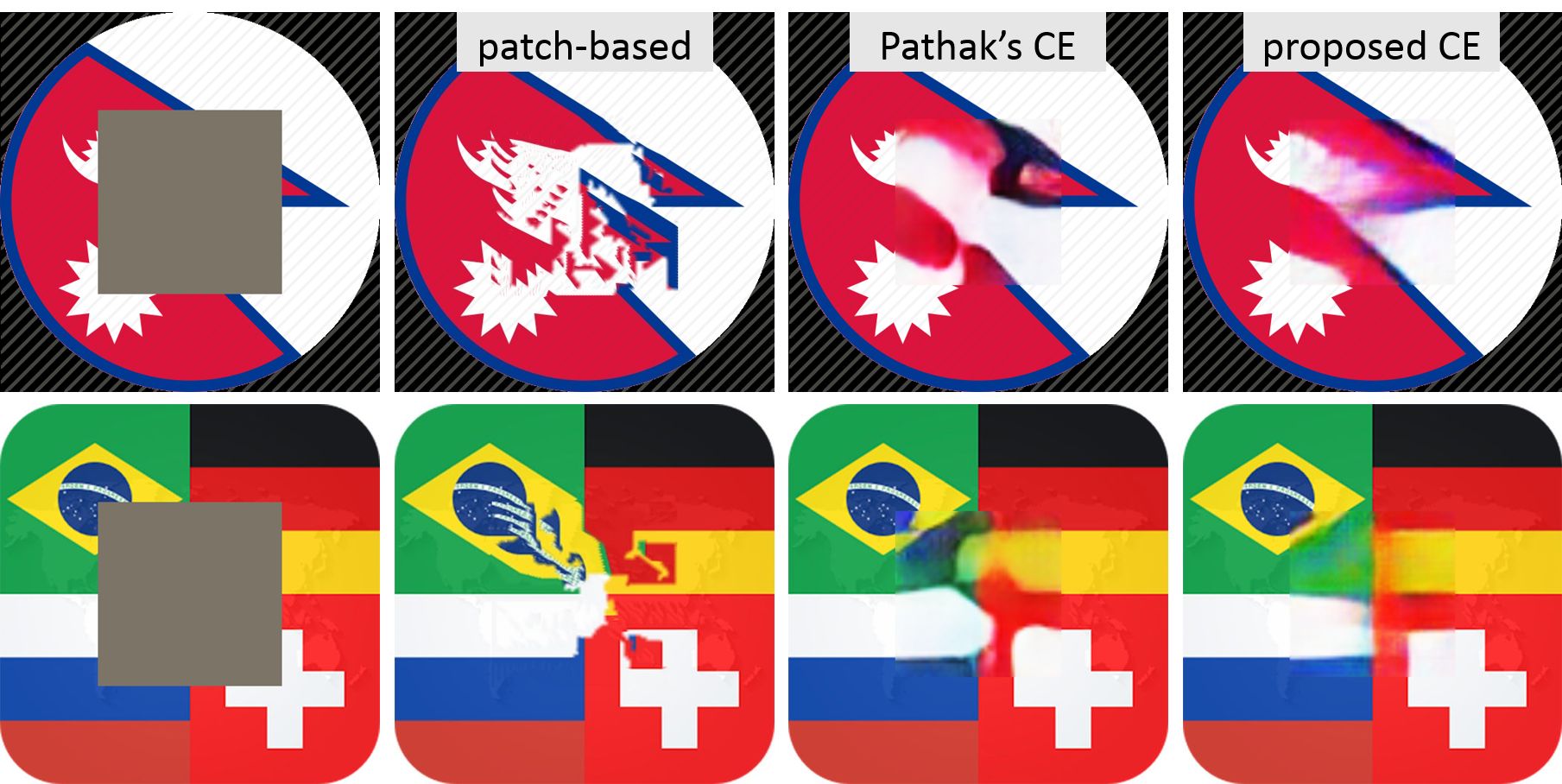}
\caption{\textbf{Completing pure structures}: State-of-the-art patch-based inpainting such as \cite{buyssens2015exemplar} fails (left); Original CE of Pathak \etal struggles as well (middle); Replacing original pixel$+$adversarial loss by a perceptual loss yields notable improvements (right).} 
\label{fig:flags}
\end{figure}

In order to assess further CE's handling of structures,
we run the code of the authors on simple graphics devoid of texture such as flags. 
Two examples of completion are shown in Fig.\ \ref{fig:flags}. Despite their apparent simplicity, such examples defeat recent patch-based methods, such as the one in \cite{buyssens2015exemplar}\footnote{Results are obtained with the G'MIC online version, \url{https://gmicol.greyc.fr/}, with default settings.} because of the length of certain structure interruptions or the absence of certain required geometric patterns in the visible part of the image. Yet, we were surprised to see that the CE was also struggling on these examples. Our interpretation is that the adversarial loss contributes way more to the texture than to the structure of the completed scene. This is also confirmed by experiments in \cite{pathak2016context} where adversarial training hardly changes the structures produced by pixel loss alone, but rather helps removing blur and gaining photo-realism. 

The key question then is whether better structures could be obtained with an alternative loss. Motivated by the successful use of pre-trained deep features to capture textures and structures at various scales in several image editing tasks, notably ``style transfer'' (whether photo-realistic \cite{luan2017deep} or not \cite{gatys2015texture}), we propose to resort to them as well for inpainting. More specifically, we consider so-called perceptual losses introduced by Johnson \etal's \cite{johnson2016perceptual} and successfully used by the authors for example-based stylization and for image super-resolution. Using such a loss to train the CE brings a benefit that is immediately visible on the previous images (Fig.\ \ref{fig:flags}, right): Stripes and other shapes are better completed, complex junctions are created where needed. These simple experiments hint toward the suitability of perceptual losses to train inpainting networks. We develop this approach in the next section with the details of the proposed context encoder.

\section{Structural inpainting}

\begin{figure}
\centering\includegraphics[width=0.95\linewidth]{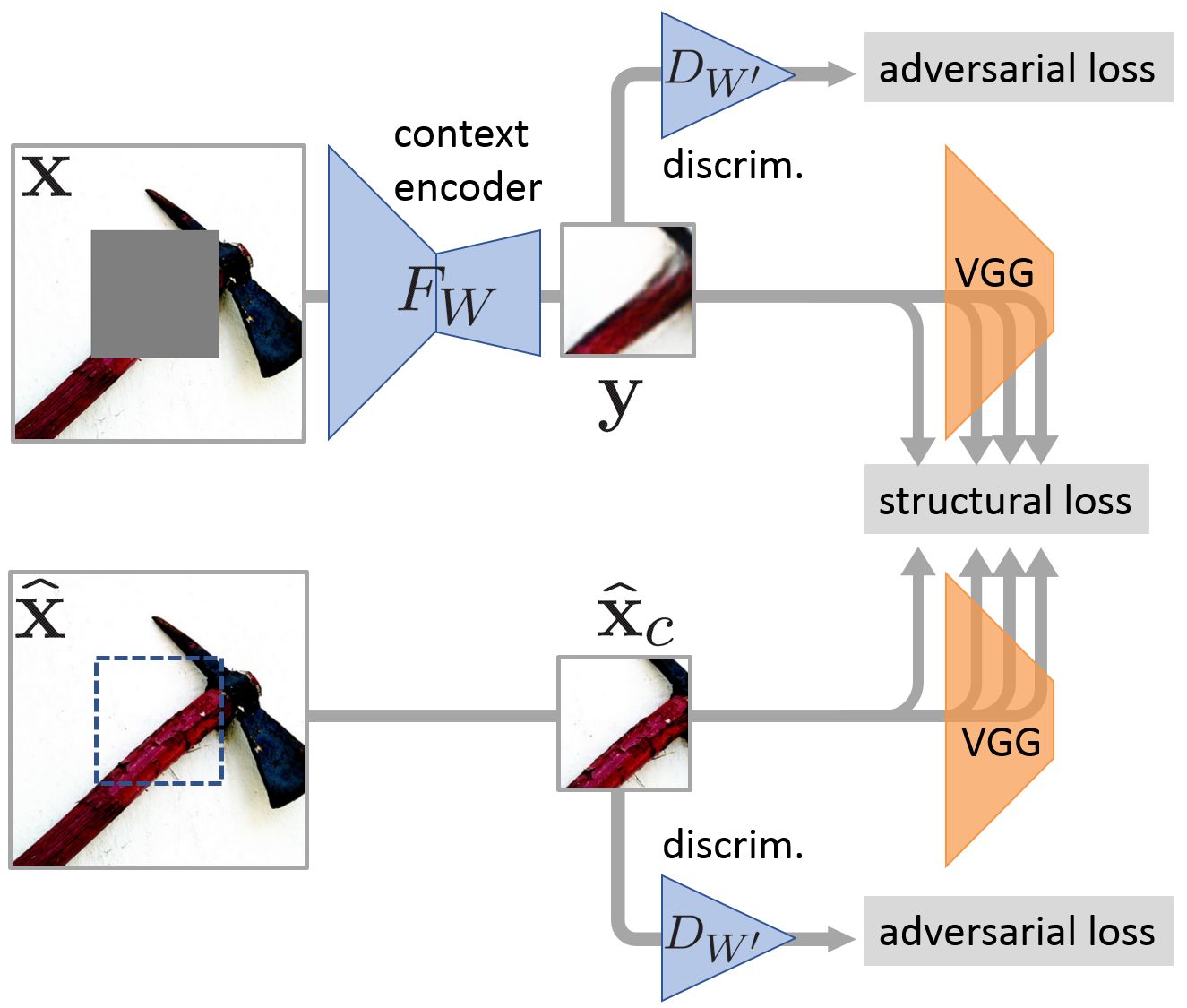}
\caption{\textbf{Proposed structural CE}: The encoder-decoder architecture of Pathak \etal is trained with a structural loss that compares the reconstructed central image part with the original one through deep features. In a second training stage, the adversarial loss of Pathak \etal is added to the total loss, with a co-trained network in charge of declaring whether an input image patch is natural or produced by the competing CE. Learnable nets are in blue, orange ones are fixed.}     
\label{fig:pipeline}
\end{figure}

The first modification we make to the original CE lies in the replacement of the pixel$+$adversarial loss by a feature-based reconstruction loss. We coin this loss ``structural'' for the reasons discussed in the previous section. The core of the proposed approach is thus as follows (Fig.\ \ref{fig:pipeline}):
\begin{itemize}
\item A convolutional encoder is combined, through a fully connected bottleneck, with a convolutional decoder of mirrored architecture but half-sized output. 
\item This encoder-decoder is trained to reconstruct the half-sized square central part of a square natural color images, this part being greyed out in the network's input. 
\item For each training image in the current training batch, both the original and the reconstructed sub-images are passed through a pre-trained convnet to extract activations at several depths. The structural loss for this training image is a combination of squared Euclidean distances in pixel space and in feature spaces, between the two images.     
\end{itemize}

More formally, let $\hat{\bx}\in\R^{M\times M\times 3}$ a complete color image and $\hat{\bx}_c\in\R^{\frac{M}{2}\times \frac{M}{2}\times 3}$ its central part. Replacing the latter by its average value (gray image patch) yields the masked image $\bx$. 
The context encoder defines a map $F_W$ parametrized by weight set $W$: 
\begin{align}
F_W:\R^{M\times M\times 3}&\to\R^{\frac{M}{2}\times \frac{M}{2}\times 3}\\
\bx&\mapsto F_W(\bx) = \by.  \nonumber
\end{align}    
The structural reconstruction loss is defined as a linear combination of pixel reconstruction error and of feature reconstruction errors computed at various convolutional layers $\ell$ of VGG-16 \cite{simonyan14very}:
\begin{equation}
\cL_{\mathrm{struct}} = \lambda_0 \cL_{\mathrm{pix}} + \sum\nolimits_{\ell} \lambda_{\ell} \cL_{\mathrm{feat},\ell},   
\label{eq:l_struct}  
\end{equation} 
with non-negative weights $\lambda_0$ and $\lambda_{\ell}$'s.
The pixel reconstruction loss reads 
\begin{equation}
\cL_{\mathrm{pix}}(\by,\hat{\bx}_c) = \| \by - \hat{\bx}_c \|_F^2. 
\end{equation}
Denoting $\phi_{\ell}(\bx)\in\R^{M_{\ell}\times M_{\ell}\times K_{\ell}}$ the stack of feature maps output by layer $\ell$ of VGG-16, the corresponding feature reconstruction loss reads:
\begin{align}
\cL_{\mathrm{feat},\ell}(\by,\hat{\bx}_c) = \| \phi_{\ell}(\by) - \phi_{\ell}(\hat{\bx}_c)\|_F^2.
\end{align}
Note that setting to zero all $\lambda_{\ell}$'s reverts to pixel reconstruction only, as experimented by Pathak \etal~\cite{pathak2016context}, and that setting all weights but $\lambda_{\tt conv2\_2}$ to 0 amounts to the loss used by Johnson \etal \cite{johnson2016perceptual} for image super-resolution.  

Given a training set $\{\hat{\bx}^{(n)}\}_{n=1}^N$, and the associated set $\{\bx^{(n)}\}_{n=1}^N$ of masked images, the empirical loss 
\begin{equation}
\frac{1}{N}\sum\nolimits_{n=1}^N \cL_{\mathrm{struct}}\big(F_W(\bx^{(n)}),\hat{\bx}_c^{(n)}\big)
\end{equation}  
is minimized w.r.t. $W$ by stochastic gradient. 

\begin{figure}[tb]
\includegraphics[width=\linewidth]{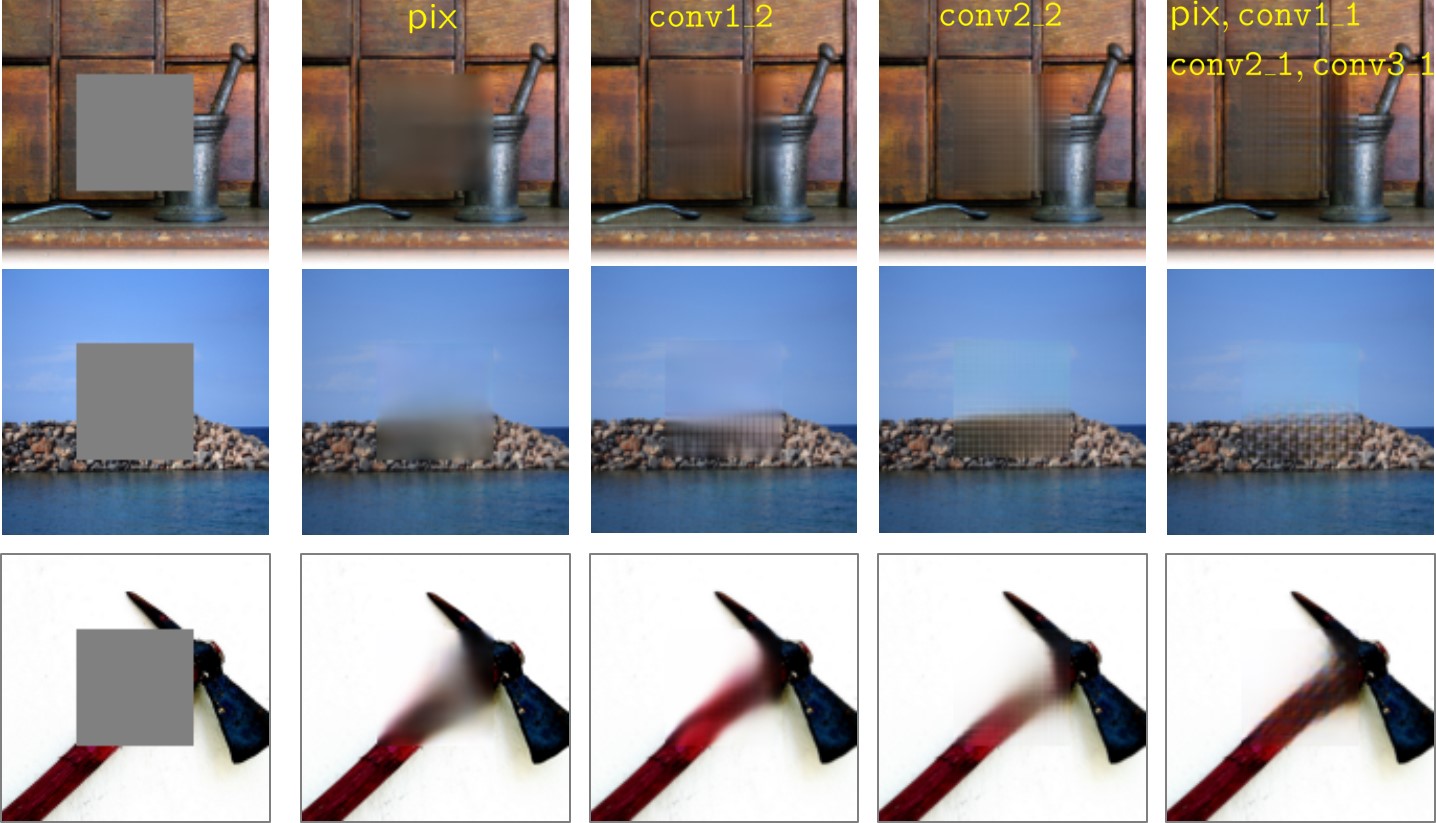}
\caption{\textbf{Different choices of structural loss} $\cL_{\mathrm{struct}}$ (Eq. \ref{eq:l_struct}): $\cL_{\mathrm{pix}}$, $\cL_{\tt{conv1\_2}}$, $\cL_{\tt{conv2\_2}}$ and $\cL_{\mathrm{pix}}+\cL_{\tt{conv1\_1}}+\cL_{\tt{conv2\_1}}+\cL_{\tt{conv3\_1}}$. 
The latter allows better, less blurry reconstruction of structures.}
\label{fig:losses}
\end{figure}

As discussed in the previous section, the adversarial loss used by Pathak \etal to train their CE helped increasing the visual quality of the reconstruction. We shall see in our experiments that this also holds when replacing the pixel reconstruction loss by the proposed structural loss. To this end, a second convolutional network $D_{W'}$, the descriminator (see Fig.\ \ref{fig:pipeline}), is simultaneously trained on central image patches to recognize synthetic ones. 
The corresponding adversarial loss for one image is defined as follows:
\begin{align}
\cL_{\mathrm{adv}}\big(F_W(\bx),\hat{\bx}_c;W'\big) = & \ln \big(D_{W'}(\hat{\bx}_c)\big) \\
& + \ln \big(1 - D_{W'}(F_W(\bx))\big).  \nonumber
\end{align} 
The joint training boils down to min-max problem ($\gamma>0$):
\begin{align}
\min_{W} \max_{W'} \frac{1}{N} \sum\nolimits_{n=1}^N  \Big[ & \cL_{\mathrm{struct}}\big(F_W(\bx^{(n)}),\hat{\bx}^{(n)}_c\big)  \label{eq:advers_train}\\  
& + \gamma \cL_{\mathrm{adv}}\big(F_W(\bx^{(n)}),\hat{\bx}^{(n)}_c;W'\big)\Big].  \nonumber
\end{align}

Regarding this self-supervised training, it is interesting to note that the targeted task, visual inpainting, is not the same as recovering the original content of the region to fill in. Actually, the most frequent motivation for inpainting is to remove an element of the original scene, hence the opposite of faithful reconstruction! Also, as discussed in the introduction, multiple equally plausible outputs are expected in a number of cases. This point might be interesting for future research. In any case, if (most) training images as such that the central region does not circumscribe completely an object, then the reconstruction training task proposed by Pathak \etal is a good proxy for inpainting. 

\section{Experiments}

\medskip\noindent{\bf Architectures}\quad Our networks are similar to those used by Pathak \etal. 
As for the encoder-decoder:\\~\\
\indent\begin{minipage}{0.95\linewidth}
{\bf Input}: Color image of size $128 \times 128 \times 3$.\\ 
{\bf Encoder}: Five convolutional layers ($4\times 4$ filters with stride 2 and ReLU) with 64, 64, 128, 256 and 512 channels respectively.\\
{\bf Bottleneck}: A fully connected layer of size 2000 (Pathak \etal use a bottleneck of size 4000, but we found that halving this size makes model and training lighter with no impact on the performance).\\
{\bf Decoder}: Four convolutional layers mirroring the last four of the encoder. In order to avoid the checker-board effect that showed up in our first experiments, we replaced the original ``deconvolutional'' design by the upsampling$+$convolution alternative proposed in \cite{zeiler2014visualizing}.\\ 
{\bf Output}: Color image of size $64\times 64\times 3$. 
\end{minipage}\\

The adversarial network takes $64 \times 64 \times 3$ inputs and is composed of four convolutional layers ($4\times 4$ filters and ReLU) with  32, 64, 128 and 256 channels respectively and a last fully connected output of size 1. It is lighter than the one in Pathak \etal, with four times fewer parameters.

\medskip\noindent{\bf Training}\quad
In order to learn a scene-agnostic, or at least general enough model, we train as Pathak \etal our CE on 1.2M images from ImageNet. 
The images in this dataset are of variable size. We resize them so that their smaller dimension is 350 and, in each epoch, we randomly crop to size $128\times 128$ each image. We use Adam stochastic solver \cite{kingma2014adam} with Nesterov momentum set to $0.5$ and with learning rates of $2\times 10^{-4}$ for the encoder-decoder and of $2 \times 10^{-5}$ for the discriminator.  

To enforce continuity at the inpainting domain's boundary, we follow Pathak \etal's technique: The masked zone is reduced to size $56\times 56$ while prediction is of size $64\times 64$. For training, the pixel reconstruction loss is scaled by 10 on pixels in the overlapping region (a band of width 4). 


\medskip\noindent{\bf Structural loss}\quad We start by investigating the exact definition of the structural loss (Eq.\ \ref{eq:l_struct}). Using only binary weights, our preliminary experiments showed that perceptual features are always beneficial, see several examples in Fig.\ \ref{fig:losses}. Whether using a single or multiple such features, structures are often better reconstructed. We empirically found that the following combination (Fig.\ \ref{fig:losses}, right-most) 
\begin{align}
\cL_{\mathrm{struct}} = \mathcal L_{\mathrm{pix}} + \mathcal L_{\mathrm{feat},\mathtt{conv1\_1}}+ \mathcal L_{\mathrm{feat},\mathtt{conv2\_1}}    
+ \mathcal L_{\mathrm{feat},\mathtt{conv3\_1}}   
\label{eq:final_l_struct}  
\end{align}
was consistently providing results of better visual quality. This is the structural loss we adopt in the next experiments. 

\begin{figure}
\centering\includegraphics[width=0.9\linewidth]{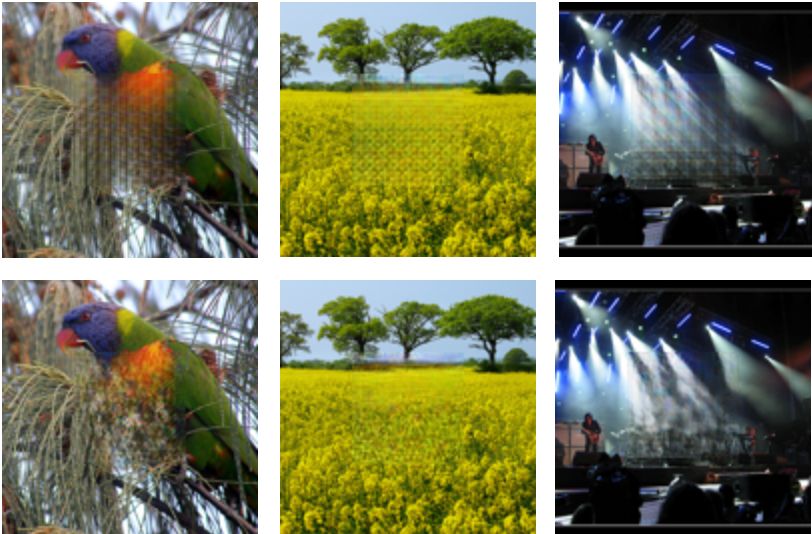}
\caption{\textbf{Grid-like artefacts with structural loss and their elimination}: When trained with our best structural loss alone, CE produces grid-like artefacts that are especially important in textured regions (top). The additional use of an adversarial loss makes these patterns disappear (bottom).}
\label{fig:pattern}
\end{figure}

\medskip\noindent{\bf Benefit of adversarial loss}\quad
While the proposed structural reconstruction loss is shown to handle better complex structures, it can be seen in Fig.\ \ref{fig:losses} that grid-like artefacts might appear in the inpainted region. These artefacts are especially pronounced in textured areas, as visible in top row of Fig.\ \ref{fig:pattern}. Adversarial training, on the other hand, is known to endow images generated by a variety of convnets with an appealing naturalness. It plays in particular a key role in Pathak \etal's CE for inpainting. For these reasons, we also resort to the adversarial training explained in Section 4. We found however that it must be handled with care. If too strong, it can become harmful to the reconstructed structures. To get only the best of it, two precautions proved useful: The discriminator is lighter, hence less powerful, than the one of Pathak \etal and it is only mobilized in a second phase of the process. This curriculum learning proceeds with 50 epochs of training with $\cL_{\mathrm{struct}}$, followed by 10 epochs of adversarial training (Eq. \ref{eq:advers_train}, with $\gamma=0.01$).  

\begin{figure}
\includegraphics[width=\linewidth]{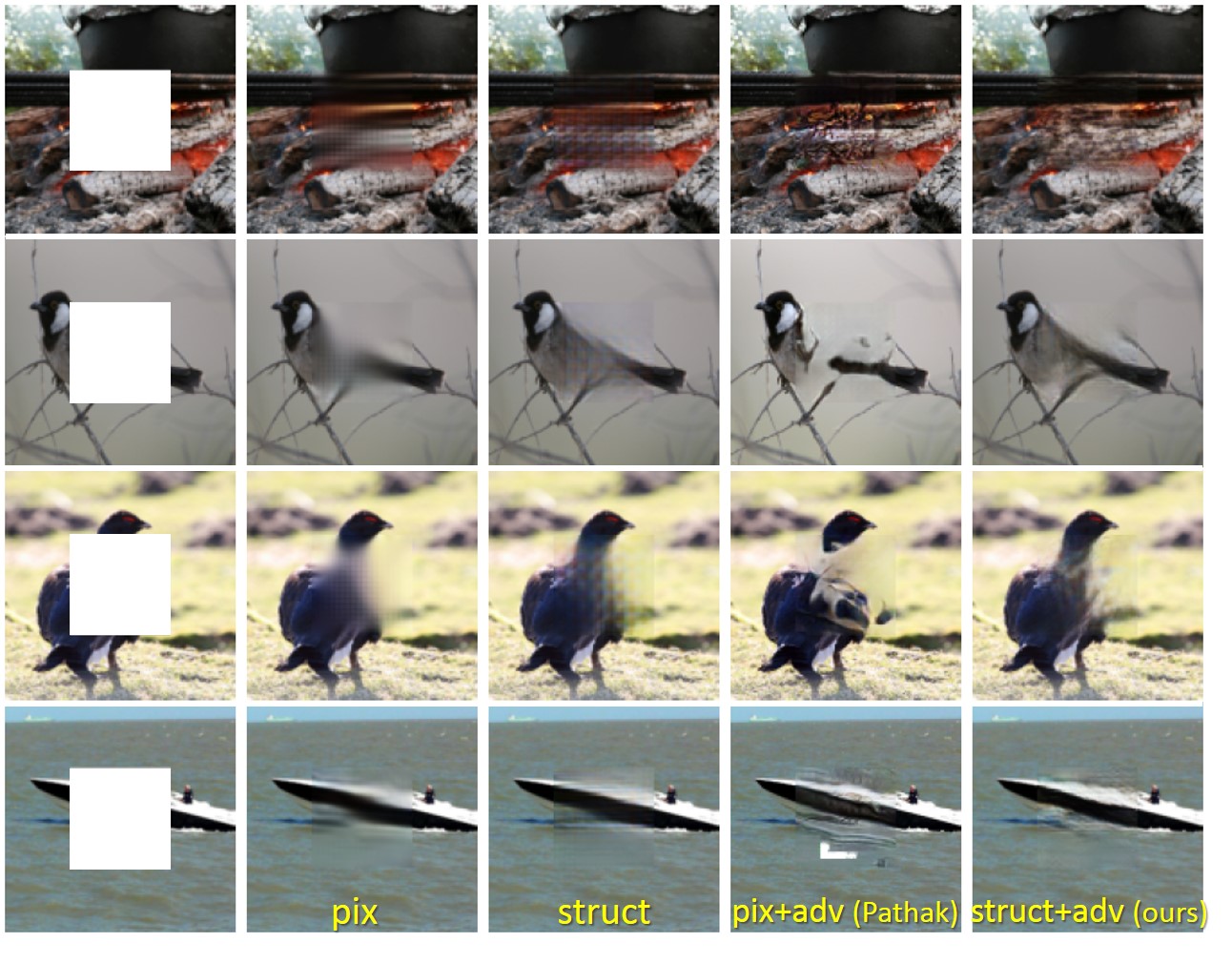}
\caption{\textbf{Different losses for inpainting context encoders}: Examples showing the respective merits of different losses to train CEs. Best visual results overall are obtained with the proposed combination of adversarial and structural losses. The latter is key to structure reconstruction while the former brings more realism.}
\label{fig:compare1}
\end{figure}

\begin{figure*}[htb]
\centering\includegraphics[width=0.95\linewidth]{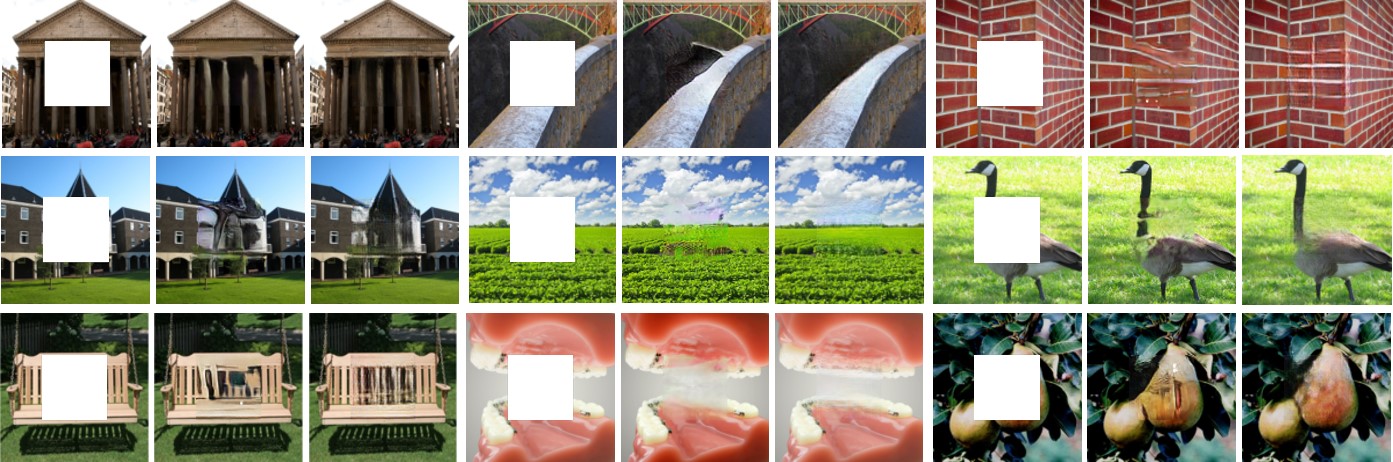}
\caption{\textbf{Comparing original and structural CEs}: For each input image (left), inpainting by Pathak \etal's CE (middle) and by the proposed CE (right). The latter offers a more plausible reconstruction of structures.}
\label{fig:final_norefine}
\end{figure*}

\begin{figure}[htb]
\centering\includegraphics[width=0.95\linewidth]{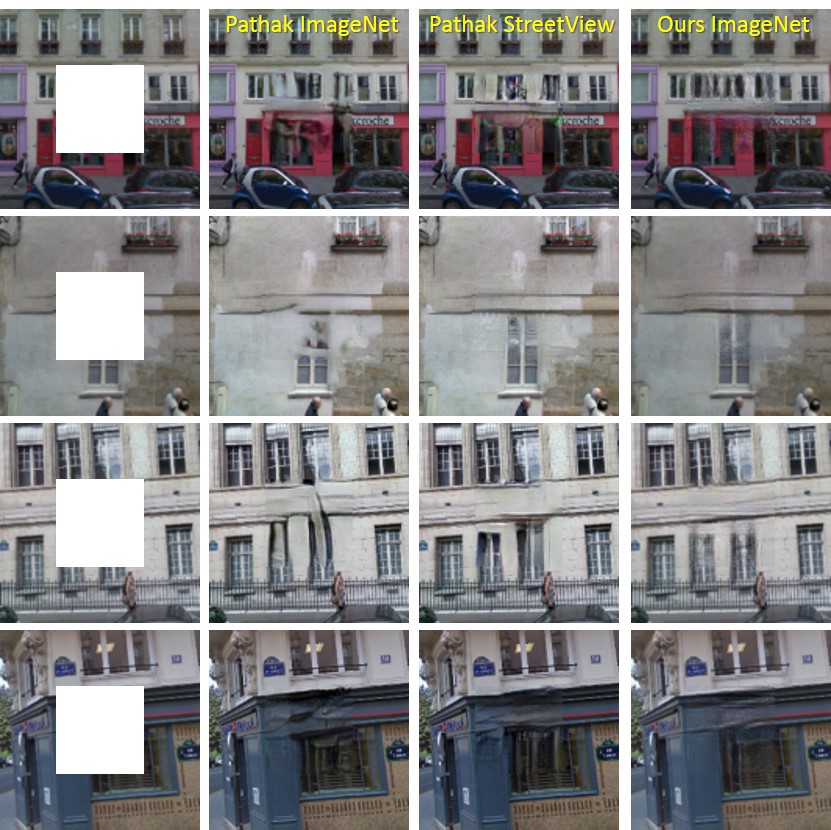}
\caption{\textbf{On urban scenes}: Despite not being specialized for this type of  very structured images, our model trained on ImageNet is on the par with Pathak \etal's CE trained on Paris Street View and substantially better than the one trained on ImageNet.}
\label{fig:streetview}
\end{figure}

\begin{figure*}[h]
\centering\includegraphics[width=0.9\linewidth]{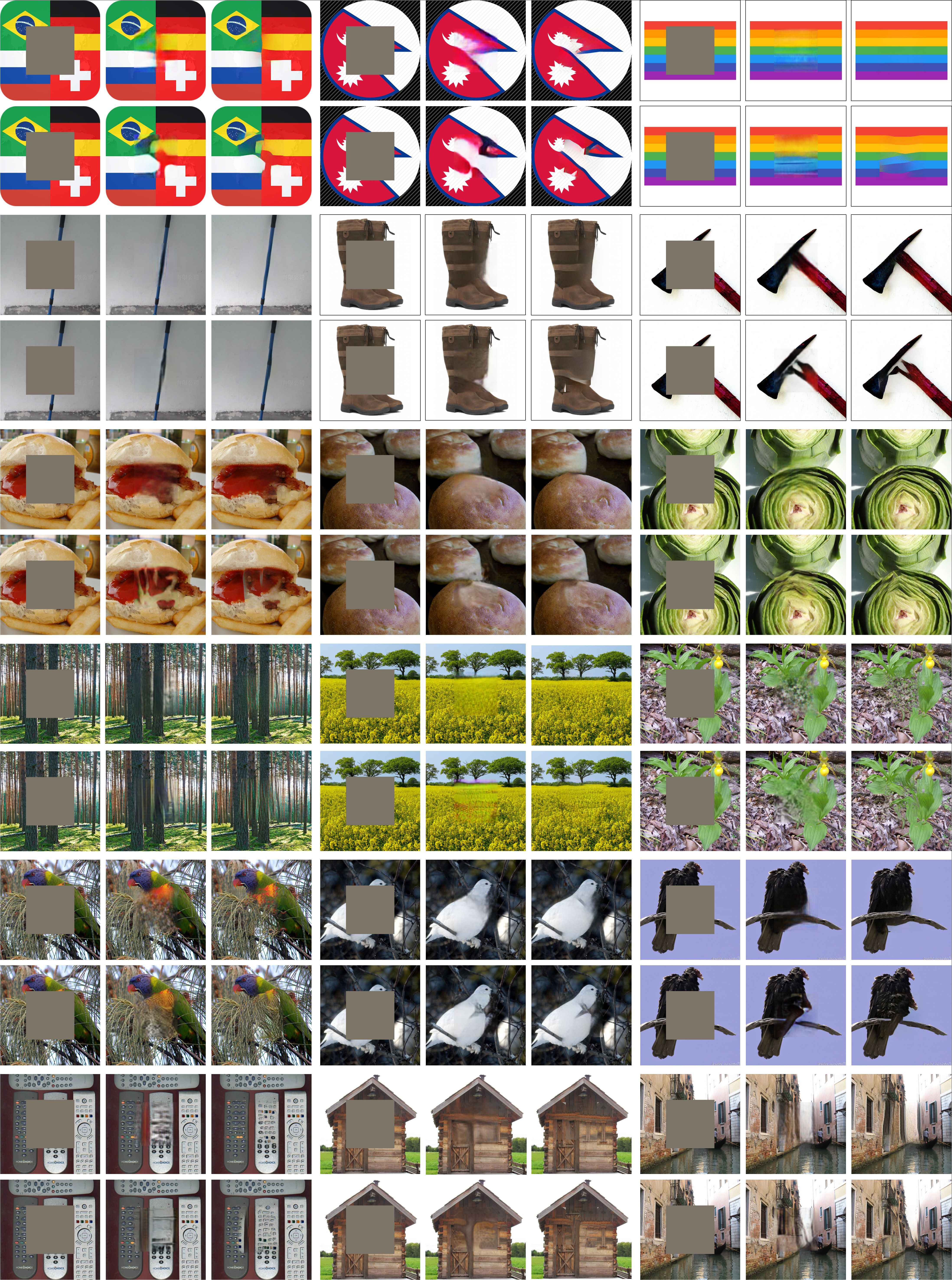}
\caption{{\bf Context encoder inpainting followed by optimization-based refinement}: For each input image, inpainting by the proposed CE, before and after optimization-based refinement (top) and same for Pathak \etal's CE (bottom). Each row contains scenes that are related in a way: Flag graphics; Simple rigid structures; Natural non-rigid objects; Multi-texture scenes; Birds on branches; More complex rigid structures.} 
\label{fig:final}
\end{figure*}

With this final phase of adversarial learning with structural loss, the proposed model is complete. The visual impact of the adversarial training is shown on several examples in Fig.\ \ref{fig:compare1}: The structural loss is responsible for the good completion of complex structures, while the addition of the adversarial loss, though costly, improves the texture and the sharpness of the inpainted regions, hence its naturalness. These examples also illustrate the qualitative improvements brought by the proposed CE in comparison to the original one. On these examples, like on the whole test collection, the reconstruction is often more plausible. Other comparisons between the two are reported in Figs.\ \ref{fig:final_norefine}, \ref{fig:streetview} and \ref{fig:final}. 


In order to highlight the structural properties of the proposed approach, we show in Fig.\ \ref{fig:streetview} result samples on street images. Although scene-agnostic, our model is doing similarly and sometimes better than Pathak \etal 's CE \textit{trained on Paris Street View} (which in turn performs poorly on other types of scene unlike ours).
This is a striking result. When trained on ImageNet like ours, the original CE does significantly worse on these highly structured scenes.

As a complement to this visual assessment, we follow \cite{pathak2016context} in reporting average pixel reconstruction errors, expressed in \% for readability and with intensities scaled in $[0,1]$, and PSNR (the larger the better) on 100 held-out ParisStreetView images:\\
\medskip
\begin{tabu}{rccc} 
		& av. $\ell_1$ error& av. $\ell_2$ error  	&  PSNR \\ [-1pt] \tabucline[1pt]{1-4}
Pathak (Paris)	& 8.37\%				& 1.63\%					& 19.57dB\\ \hline
ours (ImageNet)	& 8.07\%				& 1.49\%					& 19.89dB \\ \hline
ours (Paris)	& 7.53\%				& 1.35\%				& 20.59dB \\ \hline
\end{tabu}  
Although mostly blind to the ``visual quality'' that a good inpainting should maximise, these quantitative measures provide another hint on the impact of the proposed modifications to the original CE.


\begin{figure}
\includegraphics[width=\linewidth]{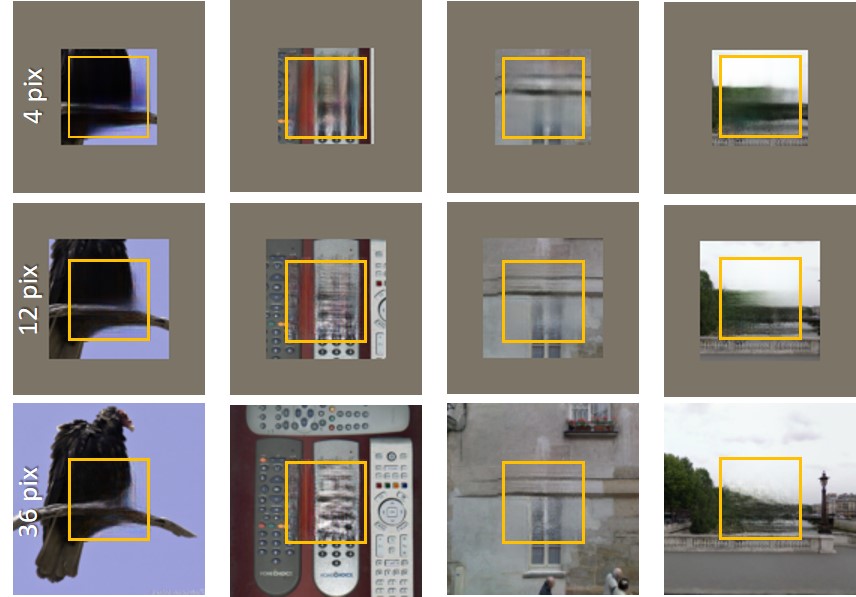}
\caption{\textbf{Varying context extent}: Structural inpainting (inside orange box) with context of 4, 12 and 36 pixels from the border respectively.
Even very restricted context allows structure completion despite lower visual quality.}
\label{fig:context}
\end{figure}

\medskip\noindent{\bf Effective context}\quad 
In this experiment, we investigate the influence of the amount of visual context on the completion. To this end we monitor the impact of masking the context beyond a certain distance in pixels to the inpainting domain. Let's start with quantitative results, expressed in terms of average pixel reconstruction errors:\vskip 5pt
\hspace{-15pt}\begin{tabu}{rcccccc} \hline
pix. & 4 & 8 & 12 & 16  & 24 & 36 \\ [-1pt] \tabucline[1pt]{1-7}
\hline
$\ell_1$ & 11.31 & 8.67 & 8.74 & 8.08 & 7.71 & 7.53 \\
\hline
$\ell_2$ & 2.11 & 1.54 & 1.54 & 1.42 & 1.38 &  1.35 \\
\hline
PSNR & 17.38 & 19.36 & 19.36 & 19.98 & 20.39  & 20.59 \\
\hline
\end{tabu} \\~\\
They demonstrate that reconstruction quality degrades gracefully with the reduction of the context and that 8 or even 4 pixels can suffice to get interesting results.
We show in Fig.\ \ref{fig:context} several examples. While the result does change, usually for the better, as the context's extent increases, it is striking that decent structure completions are possible even with as few as 4 pixels from the border known by the encoder-decoder (instead of 36 for training). Strong semantics is arguably absent from such a thin region. In our opinion, CEs contain only little object or scene-specific knowledge. Interestingly, this experiment also indicates a form of robustness. Having far less context (even worse, a partly destroyed context) yields graceful degradation. 

\medskip\noindent{\bf Results with optimization-based refinement}\quad
Despite improved structures, the image completions obtained by the proposed CE, like those obtained by the original CE, still lack visual details. As can be seen in the results reported so far, the reconstructed regions are usually blurry. This is even more acute in high-resolution images that are concerned by real-world use-cases. To circumvent this limitation of CE in the context of inpainting tasks, Yang \etal \cite{yang2016high} have recently proposed a powerful optimization-based post-processing. It builds on variational patch-based approaches (\eg, \cite{wexler2004space,demanet2003image,arias2011variational,aujol2010exemplar}) that seek a reconstruction whose patches have as good matches as possible outside the hole. Four key modifications of this paradigm are proposed by Yang \etal: An additional perceptual energy term encourages resemblance of the center $\bx_c$ of the reconstructed image with an initial prediction $\by$ by (a variant of) Pathak \etal's CE; This pre-filling also serves as initialization for the iterative solver; Total variation (TV) regularization is also included; Patches are compared in terms of VGG-based neural features, not pixels' intensities. Hence the following cost function must be minimized, in a classic coarse-to-fine multi-scale fashion:
\begin{align}
E(\bx,\psi)&  =  \alpha \sum\nolimits_{p\in\mathrm{hole}}\sum\nolimits_{\ell\in L} \big\|\phi_{\ell}(\bx,p) - \phi_{\ell}(\bx,\psi(p)) \big\|_F^2  \nonumber \\
+  \alpha' & \sum\nolimits_{\ell\in L} \big\|\phi_{\ell}(\bx_c) - \phi_{\ell}(\by) \big\|_F^2  + \beta \mathrm{TV}(\bx), 
\end{align}     
where $\bx$ is constrained to coincide with $\hat{\bx}$ outside the inpainting domain, $\psi$ is a correspondence field that maps each pixel in the hole to one outside, $\phi_{\ell}(\bx,p)$ are VGG-16 feature maps at layer $\ell$ restricted to a patch around $p$ and $L = \{{\tt conv3\_1},~{\tt conv4\_1}\}$. The minimization alternates between the two arguments, that is between correspondence map estimation given the current image reconstruction and reconstruction of central image part using the current correspondences under TV regularization and guidance from CE prediction.

In what follows we use this approach to refine the results obtained by Pathak \etal's CE and ours. 
%
%
We gather in Fig.\ \ref{fig:final} comparative results with this refinement on various types of scenes. We observe that our model improves qualitatively the results obtained by the original CE on a majority of these images and across all types of scenes, simple or complex texture-free structures, multiple textures, soft and hard contours, etc. These results demonstrate again that the use of perceptual features within the proposed reconstruction loss makes CEs better at completing structures.   

\begin{figure}
\centering\includegraphics[width=0.9\linewidth]{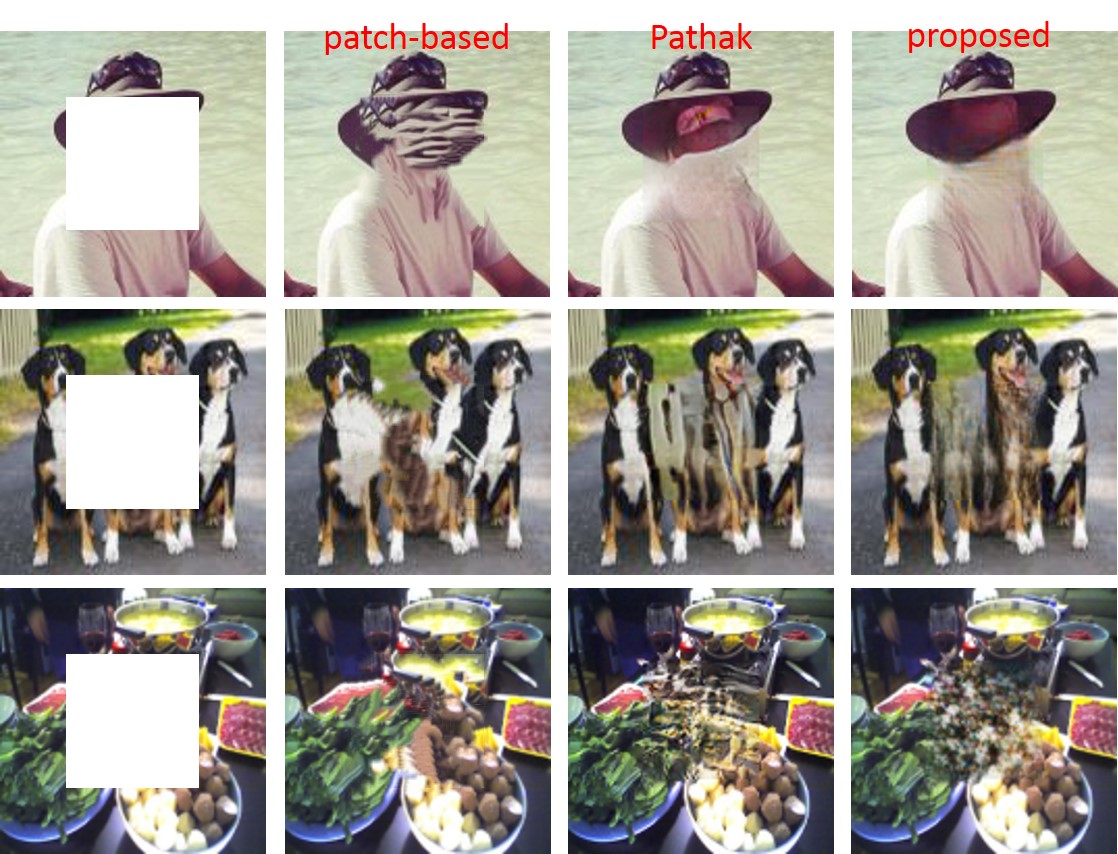}
\caption{\textbf{Failure examples}: They are many cases, like these three, where the visual/semantic complexity of the scene defeats both CEs, and patch-based methods alike, even if the proposed structural CE might fail more gracefully.}
\label{fig:failures}
\end{figure}

That being said, there is large room for improvement. Not surprisingly, the absence of explicit semantics in the model makes it unable, like other scene-agnostic approaches, to reconstruct a variety of complex scenes. In Fig. \ref{fig:failures}, we show several failure examples.

%


\medskip\noindent{\bf User study}\quad We conducted a user study involving 35 participants of various ages and occupations. The first test was run on a random set of 80 ImageNet images. For each participant, a random subset of 20 images is selected and the two reconstructions by Pathak's CE and ours are displayed in random order, alongside the incomplete image. The participants are asked to pick their preferred reconstruction, if any. Results aggregated per image (across participants) are as follows: For more than $83\%$ of the images, our reconstruction was more often preferred; for $74\%$ of the images, our reconstruction was preferred by at least $75\%$ of the participants. Fully aggregated results, across participants and images, indicate that our reconstructions are preferred $74\%$ of the time. The same test was also conducted on Paris-test images, with both CEs trained on Paris-train. Results were even more advantageous to our method in this case: $90\%$ of our inpainted images are more often preferred and our approach is preferred more that $78\%$ of the time overall. A variant of this test was also conducted with our CE \textit{not} trained on Paris but on ImageNet. We noted earlier that in this set-up, our scene-agnostic CE provided results comparable to the specialized version of Pathak's CE. The user test revealed that our CE behaves even a bit better, being preferred on $47\%$ of the images vs. $39\%$ for Pathak's (remaining $14\%$ being draws). These tests clearly indicate that our approach consistently outperforms the preceding one, across images and participants. 

In a last test, each participant is presented a random sequence of 58 images. Each image appears either in its original form or as inpainted by one of the two CEs (followed by optimization-based refinement), and the participant decides whether it looks natural or not. This is clearly a very difficult test for both inpainting methods: they both create artefacts in complex scenes and the inpainted domain is the same central region of fixed size in all examples. Yet, $51\%$ (resp. $39\%$) of the images inpainted by our method (resp. Pathak's method) were considered as natural by at least $50\%$ of participants. Interestingly, only $30\%$ of the real images received a consensus. This is clear indicator of the participants anticipating the images to be modified.

\section{Conclusion}

Context encoders can be powerful for inpainting. We have shown how their ability to complete even complex structures can be boosted by combining new reconstruction losses with careful adversarial training. Based on our experiments, we also hypothesized that semantics is playing a limited role in these feats. A deeper use of automatic scene understanding remains an open and exciting problem for inpainting in the wild. Also, in addition to relaxing current geometric constraints (inpainting a square domain), incorporating user's input in a seamless fashion would be key to the adoption of this technology in image editing systems. 



{
\small
\bibliographystyle{ieee}


\begin{thebibliography}{10}\itemsep=-1pt

\bibitem{arias2011variational}
P.~Arias, G.~Facciolo, V.~Caselles, and G.~Sapiro.
\newblock A variational framework for exemplar-based image inpainting.
\newblock {\em Int. J. Computer Vision}, 93(3):319--347, 2011.

\bibitem{aujol2010exemplar}
J.-F. Aujol, S.~Ladjal, and S.~Masnou.
\newblock Exemplar-based inpainting from a variational point of view.
\newblock {\em SIAM J. on Mathematical Analysis}, 42(3):1246--1285, 2010.

\bibitem{barnes2009patchmatch}
C.~Barnes, E.~Shechtman, A.~Finkelstein, and D.~B. Goldman.
\newblock Patchmatch: A randomized correspondence algorithm for structural
  image editing.
\newblock {\em ACM Transactions on Graphics}, 28(3):24--1, 2009.

\bibitem{bertalmio2000image}
M.~Bertalmio, G.~Sapiro, V.~Caselles, and C.~Ballester.
\newblock Image inpainting.
\newblock In {\em Proc. SIGGRAPH}, 2000.

\bibitem{bornard2002missing}
R.~Bornard, E.~Lecan, L.~Laborelli, and J.-H. Chenot.
\newblock Missing data correction in still images and image sequences.
\newblock In {\em Proc. ACM Multimedia}, 2002.

\bibitem{bugeau2010comprehensive}
A.~Bugeau, M.~Bertalm{\'\i}o, V.~Caselles, and G.~Sapiro.
\newblock A comprehensive framework for image inpainting.
\newblock {\em IEEE Trans. Image Processing}, 19(10):2634--2645, 2010.

\bibitem{burgos2015pose}
X.~Burgos-Artizzu, J.~Zepeda, F.~Le~Clerc, and P.~P{\'e}rez.
\newblock Pose and expression-coherent face recovery in the wild.
\newblock In {\em Proc. ICCVw}, 2015.

\bibitem{buyssens2015exemplar}
P.~Buyssens, M.~Daisy, D.~Tschumperl{\'e}, and O.~L{\'e}zoray.
\newblock Exemplar-based inpainting: Technical review and new heuristics for
  better geometric reconstructions.
\newblock {\em IEEE Trans. Image Processing}, 24(6):1809--1824, 2015.

\bibitem{cao2011geometrically}
F.~Cao, Y.~Gousseau, S.~Masnou, and P.~P{\'e}rez.
\newblock Geometrically guided exemplar-based inpainting.
\newblock {\em SIAM J. Imaging Sciences}, 4(4):1143--1179, 2011.

\bibitem{criminisi2003object}
A.~Criminisi, P.~P{\'e}rez, and K.~Toyama.
\newblock Object removal by exemplar-based inpainting.
\newblock In {\em Proc. Conf. Comp. Vision Pattern Rec.}, 2003.

\bibitem{criminisi2004region}
A.~Criminisi, P.~P{\'e}rez, and K.~Toyama.
\newblock Region filling and object removal by exemplar-based image inpainting.
\newblock {\em IEEE Trans. Image Processing}, 13(9):1200--1212, 2004.

\bibitem{demanet2003image}
L.~Demanet, B.~Song, and T.~Chan.
\newblock Image inpainting by correspondence maps: a deterministic approach.
\newblock {\em Applied and Computational Mathematics}, 1100(217-50):99, 2003.

\bibitem{drori2003fragment}
I.~Drori, D.~Cohen-Or, and H.~Yeshurun.
\newblock Fragment-based image completion.
\newblock 22(3):303--312, 2003.

\bibitem{efros1999texture}
A.~Efros and T.~Leung.
\newblock Texture synthesis by non-parametric sampling.
\newblock In {\em Proc. Int. Conf. Computer Vision}, 1999.

\bibitem{gatys2015texture}
L.~Gatys, A.~S. Ecker, and M.~Bethge.
\newblock Texture synthesis using convolutional neural networks.
\newblock In {\em Advances in Neural Information Processing Systems}, 2015.

\bibitem{granados2012not}
M.~Granados, J.~Tompkin, K.~Kim, O.~Grau, J.~Kautz, and C.~Theobalt.
\newblock How not to be seen—object removal from videos of crowded scenes.
\newblock In {\em Computer Graphics Forum}, 2012.

\bibitem{hays2007scene}
J.~Hays and A.~Efros.
\newblock Scene completion using millions of photographs.
\newblock 26(3):4, 2007.

\bibitem{johnson2016perceptual}
J.~Johnson, A.~Alahi, and L.~Fei-Fei.
\newblock Perceptual losses for real-time style transfer and super-resolution.
\newblock In {\em Proc. Europ. Conf. Computer Vision}, 2016.

\bibitem{kingma2014adam}
D.~Kingma and J.~Ba.
\newblock Adam: A method for stochastic optimization.
\newblock In {\em ICLR}, 2015.

\bibitem{li2017generative}
Y.~Li, S.~Liu, J.~Yang, and M.-H. Yang.
\newblock Generative face completion.
\newblock {\em arXiv preprint arXiv:1704.05838}, 2017.

\bibitem{liu2013exemplar}
Y.~Liu and V.~Caselles.
\newblock Exemplar-based image inpainting using multiscale graph cuts.
\newblock {\em IEEE Trans. Image Processing}, 22(5):1699--1711, 2013.

\bibitem{luan2017deep}
F.~Luan, S.~Paris, E.~Shechtman, and K.~Bala.
\newblock Deep photo style transfer.
\newblock In {\em Proc. Conf. Comp. Vision Pattern Rec.}, 2017.

\bibitem{masnou1998level}
S.~Masnou and J.-M. Morel.
\newblock Level lines based disocclusion.
\newblock In {\em Proc. Int. Conf. Image Processing}, 1998.

\bibitem{newson2014video}
A.~Newson, A.~Almansa, M.~Fradet, Y.~Gousseau, and P.~P{\'e}rez.
\newblock Video inpainting of complex scenes.
\newblock {\em SIAM J. on Imaging Sciences}, 7(4):1993--2019, 2014.

\bibitem{pathak2016context}
D.~Pathak, P.~Krahenbuhl, J.~Donahue, T.~Darrell, and A.~A. Efros.
\newblock Context encoders: Feature learning by inpainting.
\newblock In {\em Proc. Conf. Comp. Vision Pattern Rec.}, 2016.

\bibitem{simonyan14very}
K.~Simonyan and A.~Zisserman.
\newblock Very deep convolutional networks for large-scale image recognition.
\newblock {\em CoRR}, abs/1409.1556, 2014.

\bibitem{sulam2016large}
J.~Sulam and M.~Elad.
\newblock Large inpainting of face images with trainlets.
\newblock {\em IEEE Signal Processing Letters}, 23(12):1839--1843, 2016.

\bibitem{sun2005image}
J.~Sun, L.~Yuan, J.~Jia, and H.-Y. Shum.
\newblock Image completion with structure propagation.
\newblock {\em ACM Transactions on Graphics}, 24(3):861--868, 2005.

\bibitem{wang2007reconstruction}
Z.-M. Wang and J.-H. Tao.
\newblock Reconstruction of partially occluded face by fast recursive pca.
\newblock In {\em CISW}, 2007.

\bibitem{wexler2004space}
Y.~Wexler, E.~Shechtman, and M.~Irani.
\newblock Space-time video completion.
\newblock In {\em Proc. Conf. Comp. Vision Pattern Rec.}, 2004.

\bibitem{yang2016high}
C.~Yang, X.~Lu, Z.~Lin, E.~Shechtman, O.~Wang, and H.~Li.
\newblock High-resolution image inpainting using multi-scale neural patch
  synthesis.
\newblock In {\em Proc. Conf. Comp. Vision Pattern Rec.}, 2016.

\bibitem{yeh2017semantic}
R.~Yeh, C.~Chen, T.~Y. Lim, A.~Schwing, M.~Hasegawa-Johnson, and M.~Do.
\newblock Semantic image inpainting with deep generative models.
\newblock In {\em Proc. Conf. Comp. Vision Pattern Rec.}, 2017.

\bibitem{zeiler2014visualizing}
M.~Zeiler and R.~Fergus.
\newblock Visualizing and understanding convolutional networks.
\newblock In {\em Proc. Europ. Conf. Computer Vision}, 2014.

\end{thebibliography}
}
\end{document}